\documentclass{article}

\usepackage{microtype}
\usepackage{graphicx}
\usepackage{subfigure}
\usepackage{booktabs} % for professional tables
\newcommand{\tabitem}{~~\llap{\textbullet}~~}

\usepackage{hyperref}

\usepackage[accepted]{icml2020}

\icmltitlerunning{Impact of Legal Requirements on Explainability in Machine Learning}

\begin{document}

\twocolumn[
\icmltitle{Impact of Legal Requirements on Explainability in Machine Learning} % Title very similar to the paper title



\icmlsetsymbol{equal}{*}

\begin{icmlauthorlist}
\icmlauthor{Adrien Bibal}{equal,ml}
\icmlauthor{Michael Lognoul}{equal,legal}
\icmlauthor{Alexandre de Streel}{legal}
\icmlauthor{Beno\^{i}t Fr\'{e}nay}{ml}
\end{icmlauthorlist}

\icmlaffiliation{ml}{PReCISE, Faculty of Computer Science, NADI, University of Namur, Belgium}
\icmlaffiliation{legal}{CRIDS, Faculty of Law, NADI, University of Namur, Belgium}

\icmlcorrespondingauthor{Adrien Bibal}{adrien.bibal@unaur.be}

\icmlkeywords{Interpretability, Explainability, Machine Learning, Law}

\vskip 0.3in
]



\printAffiliationsAndNotice{\icmlEqualContribution} % otherwise use the standard text.

%\begin{abstract}
%This will be the abstract's abstract?
%\end{abstract}

\section{Legal Requirements on Explainability}
\label{sec:LegalReq}

The requirements on explainability imposed by European laws and their implications for machine learning (ML) models are not always clear. In that perspective, our research \cite{our_paper} analyzes explanation obligations imposed for private and public decision-making, and how they can be implemented by machine learning techniques.

For decisions adopted by firms or individuals, we mainly focus on requirements imposed by general European legislation applicable to all the sectors of the economy. The obligations of the General Data Protection Regulation (GDPR) (art. 13-15 and 22) as interpreted by the European Data Protection Board (EDPB) require the processors of personal data to provide ``the rationale behind or the criteria relied on in reaching the decision,'' under certain circumstances, when a fully automated decision is made (EDPB Guidelines of 3 October 2017 on Automated individual decision-making and Profiling, p. 25; see also \cite{edwards2018, wachter2017}). Consumer protection law imposes to online marketplaces to provide their consumers with ``the main parameters determining ranking [...] and the relative importance of those parameters'' (art. 6(a) of Directive 2011/83). The Online Platforms Regulation imposes very similar obligations to online intermediation services and search engines towards their professional users (art. 5 of Regulation 2019/1150).

Sectoral rules are also analyzed. For instance, financial regulators ``may require the investment firm to provide [...] a description of the nature of its algorithmic trading strategies, details of the trading parameters or limits to which the system is subject, the key compliance and risk controls that it has in place [...]. The competent authority [...] may, at any time, request further information from an investment firm about its algorithmic trading and the systems used for that trading" (art. 17(2) of Directive 2014/65 on Markets in Financial Instruments).

For decisions adopted by public authorities, two stronger requirements are studied: motivation obligations for administrations and for judges (imposed by European Convention on Human Rights). For administrative decisions, all factual and legal grounds on which the decision is based should be provided. For judicial decisions, judges have in addition to answer the arguments made by the parties in the litigation.

The objectives of those explanation requirements are twofold: first, allowing the recipients of a decision to understand it and act accordingly; second, allowing the public authority, before which a decision is contested, to exercise a meaningful effective control on the legality of the decision (European Commission White Paper of 19 February 2020 on Artificial Intelligence, p. 14).

\section{Legal Requirements and Machine Learning}
\label{sec:ML}

\begin{table*}
\centering
\small
\begin{tabular}{|p{\textwidth}|}
    \hline
    \multicolumn{1}{|c|}{\rule{0pt}{0.5cm}\textbf{Main features}\rule{0pt}{0.5cm}}\\
    \tabitem Directive 2011/83 on Consumer Rights, art. 6(a): obligation to provide the ``main parameters'' and their ``relative importance''\\
    \tabitem Regulation 2019/1150 on promoting fairness and transparency for business users of online intermediation services, art. 5: obligation to provide ``the main parameters'' and ``the relative importance of those parameters''\\
    \hline
    \multicolumn{1}{|c|}{\rule{0pt}{0.5cm}\textbf{All features}\rule{0pt}{0.5cm}}\\
    \tabitem Guidelines on automated individual decision-making and profiling: obligation to provide ``the criteria relied on in reaching the decision''\\
    \tabitem Belgian law of 4 April 2014 on insurances, art. 46: obligation to provide ``the segmentation criteria''\\
    \hline
    \multicolumn{1}{|c|}{\rule{0pt}{0.5cm}\textbf{Combination of features}\rule{0pt}{0.5cm}}\\
    Guidelines on Automated individual decision-making and Profiling: obligation to provide ``the rationale behind the decision''\\
    \hline
    \multicolumn{1}{|c|}{\rule{0pt}{0.5cm}\textbf{Whole model}\rule{0pt}{0.5cm}}\\
    Directive 2014/65 on Markets in Financial Instruments, art. 17: obligation to provide ``information [...] about its algorithmic trading and the systems used for that trading''\\
    \hline
\end{tabular}\caption{Table reproduced from \cite{our_paper} containing the legal texts used as examples in this paper.}
\label{tab:summary_examples}
\end{table*}

As explained in the previous section, legal texts do not always clearly identify the focus of the requirements. In private decision making, we identified that the explainability of four levels of machine learning entities or concepts are mentioned in legal texts \cite{our_paper}: the main features used for a decision, all features used for a decision, how the features are combined for reaching a decision and the whole model (see Table~\ref{tab:summary_examples}). 

The first and weaker level of requirements is to provide the main features used for a decision. Note that the main parameters mentioned in the legal texts refer to the features used by a ML model. While the main features used are natively provided by interpretable models such as linear models and decision trees, some works go further and provide weakly and strongly relevant features in linear models \cite{john1994,kohavi1997}. In the context of black-box models, the feature importance provided by the out-of-bag error of random forests can pass these requirements, as well as the feature importance provided through the perturbation of input feature values \cite{fisher2018}.

The second level of requirements is to provide all features involved in a decision. While providing all features used is again natively proposed by interpretable models, this requirement can be difficult to achieve when the number of features used by the model is huge. Sparsity penalties such as Lasso may be necessary to satisfy the requirement.

The third level of explainability requirements is to provide the combination of features that led to a particular decision. Again, interpretable models make it possible to check how the features have been combined to lead to a decision. In the context of black-box models, techniques like LIME \cite{ribeiro2016} have been developed to get insights on how models behave locally, i.e. for a particular decision.

Finally, the strongest requirement is to provide the whole model. In this case of strong requirement, only interpretable models can be used, as, by definition, black-box models cannot be provided (e.g. if the model is non-parametric) or understood (e.g. in the case of neural networks).

In addition to these four levels of explainability requirements for private decisions, requirements for public decisions impose two additional constraints. For administrative decisions, the legal motivation should also be provided with the decision. This means that all factual and legal grounds on which the decision is based must be provided. In the case of judicial decisions, in addition to the facts of the case and the motivation, which was already needed for administrative decisions, answers to the arguments of the parties to the litigation must also be provided. While some works try to tackle these requirements (e.g. \cite{ashley2009} explain decisions with facts only; \cite{zhong2018} introduce multi-task learning for dealing with legal articles, as well as facts; and \cite{ye2018} use sequence-to-sequence learning to propose answers to the arguments of the parties), legal requirements on the explainability of public decisions remain a challenge in machine learning, because ML algorithms are not designed to manipulate factual and legal grounds, as well as arguments, directly.

In conclusion, we call for an interdisciplinary conversation between the legal and AI research communities. In particular, legal scholars could benefit from better understanding the potential and the limitations of ML models and AI scholars from better understanding the objectives and ambiguities of the law.

\end{document}